# Text Categorization Can Enhance Domain-Agnostic Stopword Extraction

**Houcemeddine Turki, Naome A. Etori, Mohamed Ali Hadj Taieb,
Abdul-Hakeem Omotayo, Chris Chinenye Emezue, Mohamed Ben Aouicha,
Ayodele Awokoya, Falalu Ibrahim Lawan, Doreen Nixdorf**

Masakhane Research Community, Pretoria, South Africa

**Abstract**

This paper investigates the role of text categorization in streamlining stopword extraction in natural language processing (NLP), specifically focusing on nine African languages alongside French. By leveraging the MasakhaNEWS, African Stopwords Project, and MasakhaPOS datasets, our findings emphasize that text categorization effectively identifies domain-agnostic stopwords with over 80% detection success rate for most examined languages. Nevertheless, linguistic variances result in lower detection rates for certain languages. Interestingly, we find that while over 40% of stopwords are common across news categories, less than 15% are unique to a single category. Uncommon stopwords add depth to text but their classification as stopwords depends on context. Therefore combining statistical and linguistic approaches creates comprehensive stopword lists, highlighting the value of our hybrid method. This research enhances NLP for African languages and underscores the importance of text categorization in stopword extraction.

## 1. Introduction

Stopword extraction plays a pivotal role in NLP and text analysis by removing commonly occurring but semantically insignificant words, such as "the", "is", and "of" (Sarica and Luo, 2021). Hence, significantly enhances NLP models performance on various tasks such as sentiment analysis, topic modeling, and information retrieval by reducing noise, improving text understanding, ensuring consistency in analysis (Sarica and Luo, 2021). By removing stopwords, the focus shifts to meaningful content words, streamlining computational processes and improving interpretability (Dolamic and Savoy, 2009). However, it is important to tailor the list of stopwords to specific contexts, as language-specific stopword lists may be necessary, especially for languages with complex morphological structures (Dolamic and Savoy, 2009). Research in this area shows numerous advanced approaches that have evolved over time to efficiently extract stopwords from language-specific text corpora. Stopwords, which are commonly used yet carry minimal semantic value, can hinder in-depth analysis and strain computational resources.(Ferilli et al., 2014; Rani and Lobiyal, 2018).

A predominant approach to removing stopwords combines linguistic and statistical methods (Ferilli et al., 2014). Linguistic methods rely on curated lists of stopwords, encompassing common articles, conjunctions, prepositions, and other low-value words (Ferilli et al., 2014). By matching words in the text corpus against these lists, stopwords can be identified and eliminated, facilitating more meaningful analysis (Ferilli et al., 2014). While, statistical approaches employ data-driven algorithms and machine learning models to automatically pinpoint stopwords based on word frequencies and patterns within the corpus (Rani and Lobiyal, 2018; Gerlach et al., 2019). Techniques like TF-IDF and probabilistic modeling help statistically isolate and exclude stopwords, resulting in more meaningful outcomes (Rani and Lobiyal, 2018). Recent advancements in NLP and deep learning have introduced sophisticated stopwords extraction techniques. Models like BERT and GPT-3, fine-tuned for context-aware stopword identification, adapt to language nuances and enhance precision (Qiao et al., 2019). These methods strive for accuracy while accommodating diverse text corpora and languages' unique traits (Chekima and Alfred, 2016).

In this paper, we explore the potential of text categorization to simplify stopword extraction by filtering out domain-specific terms. Commonly, stopwords like articles, conjunctions, and prepositions are universally present in the text, regardless of the topic or language in focus (Gerlach et al., 2019). Their pervasive presence necessitates their removal during text analysis to ensure meaningful insights and avoid straining computational resources. We aim to validate our hypothesis for nine African languages, as well as for French, by examining the presence of stopwords in a categorized corpus of African news articles. The rest of this paper is organised as follows; We begin by detailing the language resources used in our study (Section 3). Next, we introduce our proposed approach (Section 4). We then present and discuss our findings, contextualizing them with prior research (Section 5). We conclude by summarizing our insights and suggesting avenues for future research (Section 6).

## 2. Related Work

The state-of-the-art (SOTA) stopword identification for African languages is an emerging and ongoing area of research. Given African's linguistic diversity and scarce resources, identifying stopwords in these languages poses challenges (Emezue et al., 2023). Experts are enhancing NLP methodologies for these languages by refining tailored stopword lists (Niyongabo et al., 2020), harnessing models trained on African text data (Gorro et al., 2021), and deploying rule-based tactics sensitive to linguistic subtleties (Yeshambel et al., 2022). Collaboration among linguists, NLP professionals, and African language native speakers is pivotal for progress (Emezue et al., 2023). In this context, techniques range from curated stopword lists informed by native expertise, to frequency-based tools like words frequency (Niyongabo et al., 2020), TF-IDF (Miretie and Khedkar, 2018) and Information Entropy (Asubiaro, 2013), as well as part-of-speech tagging (Ganesh et al., 2018) for grammatical insights. Additionally, machine learning models such as Naive Bayes and Recurrent Neural Networks (RNNs) are employed (Gorro et al., 2021). While rule-based strategies probe linguistic patterns, dictionary-centric methods tap into dedicated lexicons, and hybrid solutions merge various techniques, aiming for heightened stopword detection precision (Ladani and Desai, 2020).

## 3. Resources

To evaluate our hypothesis, we leveraged projects and datasets from the *Masakhane* community (Orife et al., 2020), an organization advancing African NLP. Drawing from *MasakhaNEWS* dataset (Adelani et al., 2023), *African Stopwords Project* (Emezue et al., 2023), a curated collection of stopwords, and *MasakhaPOS*, part-of-speech dataset (Bamba Dione et al., 2023) for African languages.

### 3.1. MasakhaNEWS

The MasakhaNEWS dataset (Adelani et al., 2023), addresses the scarcity of African language datasets in NLP research, by providing a news topic classification benchmark for 16 major African languages. This includes English and French as the widely-used African official languages and encompasses multiple language families, including Niger-Congo, Indo-European, English Creole, and Afro-Asiatic, representing various African regions such as East, West, etc. It involves classifying news articles into categories such as "sports", "business", "entertainment", and "politics", acting as a performance benchmark for large language models. Traditionally, NLP emphasized high-resource languages, leaving African languages underrepresented. Despite their potential, multilingual language models were limited by the absence of appropriate evaluation datasets. The MasakhaNEWS dataset provides the necessary resource for evaluating multilingual models. Masakhane community annotators used a two-stage annotation blending manual and active learning for optimal quality. The dataset is freely accessible at https://github.com/masakhane-io/masakhane-news.

### 3.2. The African Stopwords Project

The project curates stopwords for low-resource African languages, essential for NLP tasks like information retrieval. Unlike high-resource languages, African languages lack standardized stopwords, limiting NLP advancement (Emezue et al., 2023). The project seeks to curate stopwords for African languages, with progress in 10 languages thus far. The project intends to use monolingual data to discern domain-specific stopwords for African languages, with aspirations to incorporate them into NLP tools or a dedicated Python package. The dataset is currently available at https://github.com/masakhane-io/masakhanePreprocessor/tree/main/african-stopwords.

### 3.3. MasakhaPOS

MasakhaPOS (Bamba Dione et al., 2023) is a key part-of-speech (POS) dataset supporting NLP research for 20 diverse African languages. It annotates POS tags for tokens, addressing a historical gap in resources for these languages. The dataset adheres to the Universal Dependencies (UD) guidelines, ensuring consistent and comparable POS annotations across the diverse languages in MasakhaPOS. (Bamba Dione et al., 2023).
MasakhaPOS is vital for NLP research and practice and crucial for tasks like machine translation, parsing, text chunking, spell and grammar checking, and more. It boosts NLP for often-overlooked African languages due to scarce annotated datasets. Native linguistic experts ensured quality annotations. The dataset comprises training, development, and test sets, ideal for POS model training and evaluation (Bamba Dione et al., 2023). The dataset is accessible at https://github.com/masakhane-io/masakhane-pos.

## 4. Approach

We validated our hypothesis using a sample of nine African languages: two Afro-Asiatic (Hausa, Somali), six Niger–Congo (Igbo, Luganda, Kirundi, Shona, Swahili, Yoruba), and one English Creole (Nigerian Pidgin). We selected African languages based on stopword availability in MasakhaPOS and the African Stopwords Project (Table 1). For robust results, we also included

French, which is supported by MasakhaNEWS but not by MasakhaPOS or the African Stopwords Project. French stopwords were sourced from the Stopwords-ISO dataset (https://github.com/stopwords-iso), a comprehensive collection of stopwords for multiple languages.

| Language | MasakhaPOS | African Stopwords |
|---|---|---|
| Hausa (hau) | ✓ | ✓ |
| Igbo (ibo) | ✓ | ✗ |
| Luganda (lug) | ✓ | ✗ |
| Nigerian Pidgin (pcm) | ✓ | ✓ |
| Kirundi (run) | ✗ | ✓ |
| Shona (sna) | ✓ | ✗ |
| Somali (som) | ✗ | ✓ |
| Swahili (swa) | ✓ | ✓ |
| Yoruba (yor) | ✓ | ✓ |

Table 1: African languages availability in MasakhaPOS and the African Stopwords Project.

The African Stopwords Project and Stopwords-ISO feature crowdsourced stopwords, from publicly accessible lists, particularly those derived using probabilistic methods like TF-IDF (Emezue et al., 2023). We sourced the African language lists from the African Stopwords Project and used them in our study. Similarly, we obtained the French stopwords from Stopwords-ISO. As for MasakhaPOS, we extracted terms labeled with tags:*Universal Dependencies* tags: *Auxiliary Verbs* (AUX), *Pronouns* (PRON), *Coordinating Conjunctions* (CCONJ), *Subordinating Conjunctions* (SCONJ), *Determiners* (DET), and *Particles* (PART). Subsequently, we removed duplicates from the identified terms and we considered them as stopwords. We combined stopwords from MasakhaPOS, African Stopwords Project, and Stopwords-ISO into one unified list of stopwords for each studied language. then standardized them by lowercasing and removing duplicates for a concise set of stopwords for evaluation.

We then analyzed the MasakhaNEWS development set to study the distribution of these stopwords across 'news' categories. Our analysis sought to discern the distribution of stopwords among different news item categories. We first broke down the text into words, removed punctuation, and standardized all words to lowercase. This standardization facilitated consistent and accurate analysis throughout the text. After standardization, we identified unique words for each 'news' category in the MasakhaNEWS dataset. This allowed us to consistently identify unique words in each 'news' category. Finally, we calculated the presence of each stopword across MasakhaNEWS categories. This evaluation seeks to ascertain if text classification can bolster the efficiency of stopword collection.

## 5. Results and Discussion

MasakhaNEWS features a broad range of news articles with thousands of unique words across multiple languages, as shown in Table 2. These articles are sorted into distinct, non-overlapping categories, ranging from four to seven, including topics such as "business", "entertainment", "sports", and "technology" (Adelani et al., 2023). This categorization aids in testing our hypothesis on the role of text categorization in identifying domain-agnostic stopwords.

| Language | Categories | Unique words |
|---|---|---|
| French (fra) | 5 | 18290 |
| Hausa (hau) | 7 | 10495 |
| Igbo (ibo) | 6 | 8441 |
| Luganda (lug) | 5 | 8186 |
| Nigerian Pidgin (pcm) | 5 | 8057 |
| Kirundi (run) | 6 | 15363 |
| Shona (sna) | 4 | 11551 |
| Somali (som) | 7 | 14389 |
| Swahili (swa) | 7 | 18532 |
| Yoruba (yor) | 5 | 8210 |

Table 2: Statistical information about the coverage of African languages in the development set of MasakhaNEWS.

The combined use of crowdsourcing, human curation, and TF-IDF-generated word lists has proven highly effective in identifying stopwords for African languages (The African Stopwords Project) and French (Stopwords-ISO) as shown in Table 3, this collaborative approach has identified numerous stopwords for each language, underscoring the importance of both statistical techniques and human input in creating stopwords lists (Emezue et al., 2023; Rani and Lobiyal, 2018). Additionally, using MasakhaPOS (Bamba Dione et al., 2023) to automatically filter POS tags and determine stopwords based on their grammatical functions has been equally successful. This method is based on the *Universal Dependencies* POS tags. This approach, recommended by (Ferilli et al., 2014) has yielded results comparable to statistical and crowdsourcing techniques. Merging the stopwords from these two methods shows that each can detect unique stopwords, underscoring the merit of combined approaches. This supports the idea that blending multiple NLP techniques enhances stopword identification over a single method (Chekima and Alfred, 2016).

In the MasakhaNEWS dataset, most languages had a favorable stopword detection rate of over 80%, as shown in Table 4. Yet, French had a rate of 67.5%, and three Niger–Congo languages—Igbo 70.0%, Luganda 62.1%, and Yoruba 38.8%—had lower rates. The agglutinative nature of languages like Yoruba (Babarinde, 2014), Igbo (Onyenwe et al., 2019), and Luganda (Katamba, 1984), which merges stopwords with subsequent terms, might have contributed to these variances. This issue complicates stopword identification, emphasizing the need of agglutinated stopwords in stopword lists. The apostrophe (') in French (Panckhurst, 2009) might have also impacted its rate, especially since punctuation was removed during data pre-processing.

| Language | MasakhaPOS | ASP or S-ISO | Stopwords |
|---|---|---|---|
| French (fra) | N/A | 690 | **690** |
| Hausa (hau) | 90 | 321 | **329** |
| Igbo (ibo) | 70 | N/A | **70** |
| Luganda (lug) | 145 | N/A | **145** |
| Nigerian Pidgin (pcm) | 95 | 33 | **97** |
| Kirundi (run) | N/A | 59 | **59** |
| Shona (sna) | 202 | N/A | **202** |
| Somali (som) | N/A | 30 | **30** |
| Swahili (swa) | 97 | 103 | **156** |
| Yoruba (yor) | 122 | 60 | **160** |

Table 3: Statistical information about the identification of stopwords from MasakhaPOS and the African Stopwords Project (ASP, for African languages) or Stopwords-ISO (S-ISO, for French).

| Language | Found Stopwords | Available in *N* Categories | | | | | | |
|---|---|---|---|---|---|---|---|---|
| | | 1 | 2 | 3 | 4 | 5 | 6 | 7 |
| French (fra) | 466 (67.5%) | 69 | 43 | 61 | 64 | 229 | N/A | N/A |
| Hausa (hau) | 319 (97.0%) | 16 | 20 | 19 | 28 | 38 | 52 | 146 |
| Igbo (ibo) | 49 (70.0%) | 5 | 4 | 5 | 2 | 6 | 27 | N/A |
| Luganda (lug) | 90 (62.1%) | 11 | 14 | 11 | 16 | 38 | N/A | N/A |
| Nigerian Pidgin (pcm) | 84 (86.6%) | 2 | 7 | 11 | 11 | 53 | N/A | N/A |
| Kirundi (run) | 55 (93.2%) | 3 | 0 | 0 | 1 | 6 | 45 | N/A |
| Shona (sna) | 175 (86.6%) | 21 | 30 | 31 | 93 | N/A | N/A | N/A |
| Somali (som) | 25 (83.3%) | 0 | 1 | 1 | 0 | 0 | 0 | 23 |
| Swahili (swa) | 151 (96.8%) | 1 | 3 | 5 | 3 | 11 | 59 | 69 |
| Yoruba (yor) | 62 (38.8%) | 3 | 3 | 3 | 4 | 49 | N/A | N/A |

Table 4: Distribution of identified stopwords based on their presence across MasakhaNEWS categories. *N* corresponds to the number of categories where keywords exist.

| Language | Uncommon stopwords |
|---|---|
| Hausa (hau) | **dana** (son), *dari* (one hundred), **lalle** (certainly), yayinda (while), shima (too), *milyan* (million), **guji** (avoid), *balle* (let alone), kanka (yourself), basa (they are not), namu (ours), sune (they are), **kwarai** (absolutely or extremely), wadancan (those), **daman** (right side), *daukacin* (all) |
| Igbo (ibo) | i (you), o (user), *ozo* (again), **ihi** (reason), *imirikiti* (many) |
| Luganda (lug) | teyali (it was not), **ebyali** (he was eating), ekyali (which was), alabika (appears), byali (were), **okwaliwo** (which occurred), **yalina** (it was raining), **kaali** (cabbage), *egisinga* (most of them), bakyali (they are still), kyayo (its own) |
| Nigerian Pidgin (pcm) | sey (as if), non (on point) |
| Kirundi (run) | kw (on, at, in, and from), nk (seems like), |
| Shona (sna) | ry (there), **vashaye** (miss them), haafanirwe (he should not), inove (it is), *uchiri* (you are still), racho (its), ndave (I have been), **vaifanirwa** (they deserved it), *dzinogara* (they last), **panga** (sword), *dzinenge* (almost), dzainge (were), raro (its), sezviri (as it is), vavari (who they are), chavari (what they are), mumwe (one), neku (and), **haro** (necessarily), hunogona (it can), vangadai (they would have), **dzimwe** (others) |
| Somali (som) | N/A |
| Swahili (swa) | yasio (not) |
| Yoruba (yor) | í (it), kì (do not), é (yes) |

Table 5: List of stopwords featured in a unique MasakhaNEWS category for the nine considered African languages (with meaning).

The analysis shows over 40% of considered stopwords appear across all MasakhaNEWS categories, with high rates in Somali 92.0%, Kirundi 81.8%, Yoruba 79.0%, and Nigerian Pidgin 63.1% (Table 4). Less than 15% are unique to one category, with low instances in Somali 0.0%, Swahili 0.6%, Nigerian Pidgin 2.4%, and Yoruba 4.8%.

Uncommon stopwords, highlighted in bold in Table 5, add depth to texts with nouns, verbs, and adverbs, often acting as verbal cues (Treisman, 1964), especially in narrative-driven African languages (Abdi, 2009). However, not all are definitively stopwords. But context matters; terms like numerals, frequency-related adverbs and adjectives, time, or color (italicized in Table 5) are not always stopwords. Some words carry meaning in specific contexts and structures (Jóhannsdóttir, 2007; Keenan and Stavi, 1986). Thus, text categorization efficiently identifies domain-agnostic stopwords.

## 6. Conclusion

In conclusion, our study on text categorization's influence on stopword extraction in the MasakhaNEWS dataset, covering languages like French, Hausa, and Yoruba, offers key insights. Using linguistic methods, statistics, and Masakhane community experts, we pinpointed general stopwords. We found that text categorization reliably identifies common stopwords across news categories. However, detection rates vary, especially in languages with intricate linguistic characteristics, highlighting the need for language-specific considerations in stopword identification.

Furthermore, our analysis identified unique stopwords that add depth and meaning to news categories often including nouns, verbs, and adverbs. This emphasizes the importance of context in stopword extraction. Future research will focus on expanding language coverage, use of context-aware techniques for stopwords extraction, utilizing multilingual models, hybrid techniques, addressing challenges with agglutinative languages, and standardized metrics. Moreover, we see the need for domain-specific stopwords, tailored tools, cross-language methods, collaboration with linguistic experts, and integrating ethical aspects.

## 7. Conflict of interest

## 8. Acknowledgements

We thank David Adelani (University College London, United Kingdom) and Akintunde Oladipo (University of Waterloo, Canada) for providing useful comments and discussion regarding this work. Certain portions of this work have undergone language proofreading and editing with the assistance of *ChatGPT*, a robust chatbot powered by OpenAI's advanced language model.

## 9. Data Availability

Source data links are provided throughout this manuscript. The meaning of the uncommon stopwords in Table 5 are collected from Google Translate (https://translate.google.ca), Hausa Dictionary (https://hausadictionary.com), Kirundi Study and Dictionary (https://www.matana.de/index1.php), Glosbe (https://glosbe.com/), Naija Lingo (http://www.naijalingo.com/), and the Masakhane Community.

## 10. Source code

The source code is available at https://github.com/masakhane-io/masakhanePreprocessor/blob/main/african-stopwords/Masakhane_Stopword_Extraction_and_Validation.ipynb for reproducibility purposes.